\documentclass[letterpaper, 10 pt, conference]{ieeeconf}  

\IEEEoverridecommandlockouts                              

\usepackage{amsmath,amsfonts,bm}

\def\eqref#1{equation~\ref{#1}}

\def\1{\bm{1}}

\def\vtheta{{\bm{\theta}}}
\def\va{{\bm{a}}}

\def\vd{{\bm{d}}}
\def\ve{{\bm{e}}}

\def\vo{{\bm{o}}}

\def\vs{{\bm{s}}}

\def\vz{{\bm{z}}}

\def\vtheta{{\bm{\theta}}}
\def\vphi{{\bm{\phi}}}

\DeclareMathAlphabet{\mathsfit}{\encodingdefault}{\sfdefault}{m}{sl}
\SetMathAlphabet{\mathsfit}{bold}{\encodingdefault}{\sfdefault}{bx}{n}

\def\gA{{\mathcal{A}}}

\def\gO{{\mathcal{O}}}

\def\gS{{\mathcal{S}}}
\def\gT{{\mathcal{T}}}

\def\gW{{\mathcal{W}}}

\def\gZ{{\mathcal{Z}}}

\def\sR{{\mathbb{R}}}

\newcommand{\E}{\mathbb{E}}

\usepackage[utf8]{inputenc} 
\usepackage[T1]{fontenc}    
\usepackage{hyperref}       
\usepackage{url}            
\usepackage{booktabs}       
\usepackage{nicefrac}       
\usepackage{microtype}      
\usepackage{cite}
\usepackage{tabularx}
\usepackage[table]{xcolor}
\usepackage{colortbl}
\usepackage{graphicx}
\usepackage{pifont}
\usepackage{tikz}
\usepackage{array}
\usepackage{wrapfig}
\usepackage{multirow}
\usepackage{threeparttable}
\usepackage{algorithm}
\usepackage{algpseudocode}
\usepackage{float}
\usetikzlibrary{positioning, arrows.meta, fit, backgrounds}
\newcommand{\cmark}{\ding{51}}
\newcommand{\xmark}{\ding{55}}

\usepackage{etoolbox}
\makeatletter
\patchcmd{\@makecaption}{\scshape}{}{}{}
\makeatother

\title{\LARGE \bf
Benchmarking World Models \\ for Continual Learning on Compositional Tasks
}

\author{Haoyu Zhou*, Joe Watson, Anson Lei, and Ingmar Posner%
\thanks{All authors are with the Applied Artificial Intelligence (A2I) Lab, Department of Engineering Science, University of Oxford.
       {\tt\small \{haoyu, joewatson, anson, ingmar\}@robots.ox.ac.uk}}%
\thanks{* Corresponding author}%
}

\begin{document}

\maketitle
\thispagestyle{empty}
\pagestyle{empty}

\begin{abstract}
A desirable property of a world model is the ability to learn continually across tasks, adapting to new environments without forgetting what the agent has already learnt.
In particular, the ability to retain and reuse knowledge obtained from prior experiences underpins an agent's ability to efficiently adapt to novel environments, as the dynamics of the physical world can often be described in recurring mechanisms. 
However, the world model's measure of adaptation entangles two abilities: the speed and capacity to learn unseen tasks, and the reuse of knowledge already acquired, since incoming tasks carry novel content alongside what recurs.
In order to isolate knowledge reuse from prior experiences, we propose a compositional continual learning benchmark for world models in robot manipulation.
Specifically, we design each task curriculum with \textit{compositional} tasks that combine aspects of the tasks seen in the sequence.
We further factorise this composition along the axes of action and perception to better understand how different input modalities bottleneck knowledge reuse.
We evaluate state-of-the-art world models under canonical continual learning methods, alongside a \textit{modular} world model whose dynamics backbone contains explicitly reusable components.
Results show that modularity balances reuse against forgetting better than conventional methods, but none solve the problem fully, leaving clear room for continual world models built to reuse without forgetting.
More details are available on our \href{https://object814.github.io/Compositional-Continual-Learning/}{project website}.
\end{abstract}

\section{Introduction}
\label{sec:introduction}

\begin{figure}[t!]
    \centering
    \includegraphics[trim={5pt 408pt 128pt 30pt}, clip, width=\linewidth]{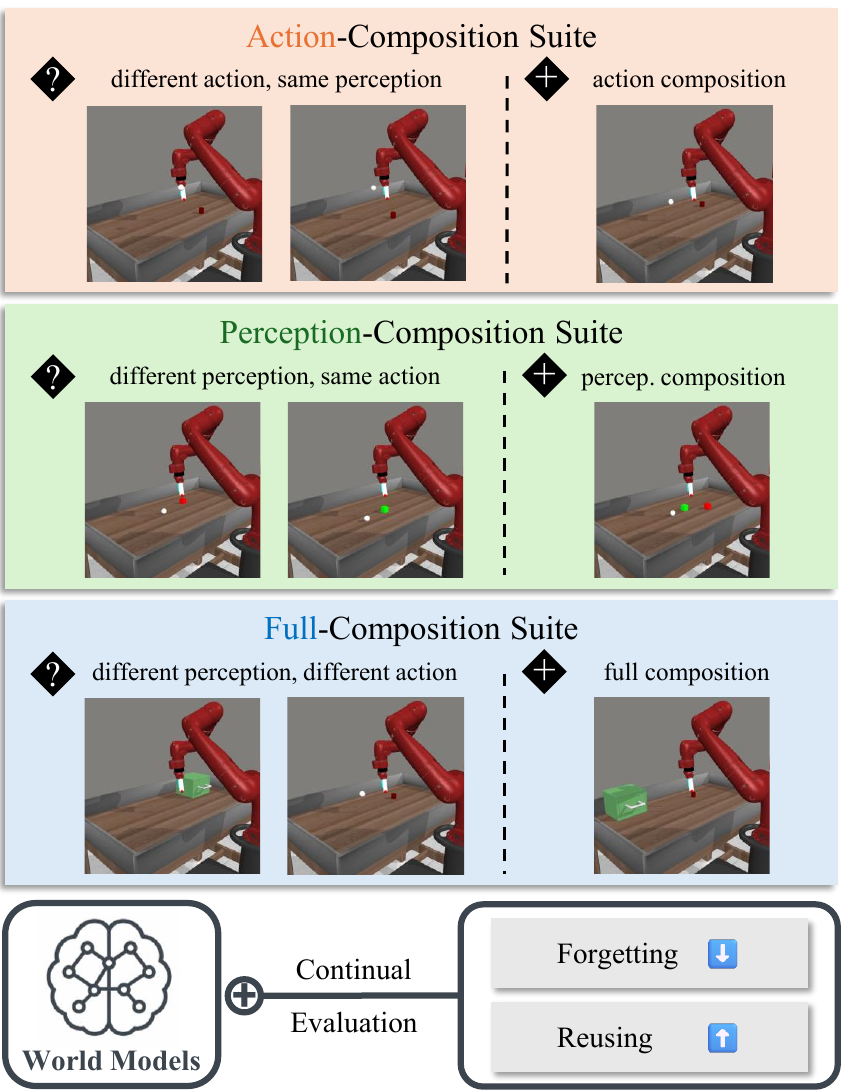}
    \vspace{-15pt}
    \caption{\textbf{Overview of our compositional continual learning benchmark for world models in robot manipulation.}
    \textit{Top}: Each task suite presents a sequence of primitive manipulation tasks, ending with a composition task that recombines them across action, perception, or full composition axes.
    \textit{Bottom}: We evaluate the continual learning ability of world models to efficiently adapt to the composition tasks through knowledge reuse, without forgetting previously learnt primitives along the sequence.}
    \label{fig:teaser}
    \vspace{-15pt}
\end{figure}

\begin{table*}[t!]
\vspace*{6pt}
\centering
\caption{Comparison of robot manipulation benchmark designs. \textbf{Task Suite Structure} groups related tasks into suites, \textbf{Compositional Structure} exposes the model with primitive tasks then test compositional generalisation, \textbf{Continual Learning Protocol} forces the models to learn in a curriculum and evaluate forgetting while adaptation happens, \textbf{Dense, Monolithic Reward} enables RL benchmarking, and \textbf{Expert Demo/Policy} provide expert demonstration for offline training. While many benchmarks look at different bits, ours evaluates them comprehensively.}
\label{tab:benchmark_table}
\vspace{-5pt}
\colorlet{checkgreen}{green!10}
\newcommand{\cmarkc}{\cellcolor{checkgreen}\cmark}
\scriptsize
\setlength{\tabcolsep}{6pt}
\renewcommand{\arraystretch}{0.95}
\begin{tabular*}{\textwidth}{@{}l@{\extracolsep{\fill}}ccccc@{}}
\toprule
 & \textbf{Task Suite} & \textbf{Compositional} & \textbf{Continual Learning} & \textbf{Dense, Monolithic} & \textbf{Expert} \\
 & \textbf{Structure} & \textbf{Structure} & \textbf{Protocol} & \textbf{Reward} & \textbf{Demo/Policy} \\
\midrule
BEHAVIOR-1K         & \xmark  & \xmark  & \xmark  & \xmark  & \cmarkc \\
CALVIN              & \cmarkc & \cmarkc  & \xmark  & \xmark  & \cmarkc \\
CausalWorld         & \cmarkc & \xmark  & \xmark  & \cmarkc & \xmark  \\
Colosseum           & \xmark  & \xmark  & \xmark  & \xmark  & \cmarkc \\
CompoSuite          & \xmark  & \cmarkc & \xmark  & \cmarkc & \cmarkc  \\
Continual World     & \cmarkc & \xmark  & \cmarkc & \cmarkc & \xmark  \\
CRIL                & \cmarkc & \xmark  & \cmarkc & \xmark  & \cmarkc \\
Franka Kitchen      & \xmark  & \cmarkc & \xmark  & \xmark  & \cmarkc \\
iManip              & \cmarkc & \xmark  & \cmarkc & \xmark  & \cmarkc \\
LIBERO              & \cmarkc & \cmarkc & \cmarkc & \xmark  & \cmarkc \\
ManiSkill3          & \cmarkc & \xmark  & \xmark  & \cmarkc & \cmarkc \\
MetaWorld           & \cmarkc  & \xmark  & \xmark  & \cmarkc & \cmarkc \\
NBAgent             & \cmarkc & \xmark  & \cmarkc & \xmark  & \cmarkc \\
RLBench             & \cmarkc & \xmark  & \xmark  & \xmark  & \cmarkc \\
RoboCasa365         & \cmarkc & \cmarkc & \cmarkc & \xmark  & \cmarkc \\
RoboCerebra         & \cmarkc & \cmarkc & \xmark  & \xmark  & \cmarkc \\
RoboMimic           & \xmark  & \xmark  & \xmark  & \cmarkc & \cmarkc \\
VIMA-Bench          & \cmarkc & \cmarkc & \xmark  & \xmark  & \cmarkc \\
VLMbench            & \xmark  & \cmarkc & \xmark  & \xmark  & \cmarkc \\
\midrule
\textbf{Ours}        & \cmarkc & \cmarkc & \cmarkc & \cmarkc & \cmarkc \\
\bottomrule
\end{tabular*}
\vspace{-15pt}
\end{table*}

Agents should be able to solve a sequence of related tasks efficiently by using and reusing their acquired knowledge.
Methods using world models are a natural solution, as the learnt dynamics model is a persistent component across tasks.
In this work, we propose a benchmark for evaluating continual reinforcement learning with world models, specifically looking at robot manipulation due to its compositional nature.

Existing continual learning benchmarks in robot manipulation~\cite{liu2023libero,wolczyk2021continual} investigate how well models
adapt to new tasks without forgetting previous ones. Their measure of adaptation, however, usually entangles the raw speed of learning new interactions, and the ability to reuse previous knowledge.
This entanglement arises because every task arriving in their sequences carries novel, previously unseen content alongside the objects and actions that recur.
In the context of world models, we argue that \textit{reuse} is the main route to efficient adaptation, as the dynamics of the environment can often be explained by a small set of reusable mechanisms~\cite{scholkopf2021toward,posner2026observation,momennejad2026compositional}. 

In this light, we propose a benchmark that isolates the reusability of world models in model-based reinforcement learning.
We design task sequences that end with a \textit{composition} task built by recombining the primitive tasks that precede it, examining how well world models are able to reuse without forgetting.
We further factorise this composition along the two inputs a world model is conditioned on, namely \emph{action} and \emph{perception}, into separate task suites.
The results therefore reveal not only whether a model reuses, but also along which input modality that reuse succeeds or fails.

We evaluate state-of-the-art world models paired with conventional continual learning methods on our benchmark to see how far they carry world models towards compositional continual learning. Additionally, we look into a \textit{modular} world model, whose dynamics are a mixture of experts rather than a monolithic network. Such an explicitly factorised model is a direct architectural expression of our view, since its modules are candidate reusable components that mirror the compositional structure of the environment.
Whether modular world models enable compositional continual learning is what our benchmark asks.
We also introduce a principled way to separate world models into a task-agnostic backbone and task-specific heads, applying continual learning only to the part meant to capture reusable mechanisms.
In summary, our key contributions are:
\begin{itemize}
    \item \textbf{A compositional continual learning benchmark.}
    We build a benchmark for world models in robot manipulation, where each task sequence ends with a task constructed by recombining previous primitives, and is factorised along \emph{action} and \emph{perception}, so that forward transfer isolates reuse and localises where it fails.
    \item \textbf{A principled separation of world models for continual learning.} We distinguish the task-agnostic backbone that must be learnt continually from the task-specific heads that need not be, and apply continual learning methods only to the former.
    \item \textbf{An evaluation of state-of-the-art world models.} We evaluate monolithic world models under canonical continual learning methods to see how far they carry world models towards compositional continual learning, alongside a continual variant we construct on a modular world model to see what modularity adds.
\end{itemize}


\section{Related Work}
\label{sec:relatedworks}

\paragraph{General-purpose robot manipulation benchmarks}
A first wave of manipulation benchmarks has emphasised broad task coverage, visual realism, and policy generalisation in single or multi-task regimes. Several benchmarks provide diverse manipulation suites with expert demonstrations or meta-learning structure~\cite{james2020rlbench,yu2020meta}, while others scale robot learning through larger asset libraries, multiple embodiments, and more realistic household scenes~\cite{gu2023maniskill2,tao2024maniskill3,nasiriany2024robocasa,li2023behavior}. A complementary line focuses on demonstration-driven evaluation, offline learning, and long-horizon manipulation, from structured human demonstration datasets to kitchen-style and household tasks~\cite{fu2020d4rl,han2026robocerebra}.
Together they provide rich task diversity, abundant demonstrations, and vivid rendering, but target single-task or multi-task learning rather than the sequential task arrival a general-purpose robot meets over its lifetime.

\paragraph{Continual learning robot manipulation benchmarks}
A more closely related cluster of benchmarking works directly evaluates the continual learning performance in robot manipulation settings. Continual World~\cite{wolczyk2021continual} arranges sequences of Meta-World~\cite{yu2020meta} tasks and evaluates continual RL agents through forward-transfer and forgetting metrics, but its sequences are built from monolithic tasks, which does not align with evaluating the compositional aspect of continual learning. LIBERO~\cite{liu2023libero} extends continual robot learning with task suites organised around spatial, object, goal, and mixed distribution shifts. However, these shifts are pairwise rather than cumulative: each task varies relative to its neighbours, and no task is defined as a recombination of everything that precedes it. A model can score well by adapting to the most recent shift alone, without ever reusing its full history.

\paragraph{Composition robot manipulation benchmarks}
A third cluster targets compositional or causal generalisation explicitly. 
CompoSuite~\cite{mendez2022composuite} factorises 256 manipulation tasks along robot, object, objective and obstacle axes, exposing shared components across tasks; VIMA-Bench~\cite{jiang2023vima} formalises multimodal tasks under a four-level generalisation protocol; CALVIN~\cite{mees2022calvin} composes long-horizon language-conditioned skills across four environments; THE COLOSSEUM~\cite{pumacay2024colosseum} systematically perturbs RLBench tasks along fourteen environmental axes; and VLMbench~\cite{zheng2022vlmbench} provides a compositional vision-and-language manipulation suite. CausalWorld~\cite{ahmed2020causalworld} sits adjacent to this line, enabling fine-grained interventions over causal factors such as mass, colour, friction to study transfer.
These benchmarks make the compositional structure of manipulation explicit, but evaluate it only under a static training regime. The agent is given the full task distribution up front, with no continual arrival nor measurement of whether previously learnt content is retained while reusing.
We summarise this comparison in Table~\ref{tab:benchmark_table}.


\section{Compositional Continual Learning Benchmark for World Models}
\label{sec:method}

\subsection{Problem setting}
\label{sec:problem}

We consider a curriculum of $T$ tasks arriving in sequence, where each task is a partially observed Markov decision process
$\gT_t = (\gS_t, \gA, \gO, p_t, r_t, \gamma)$ with underlying states $\vs_h \in \gS_t$, actions $\va_h\in\gA$, observations $\vo_h\in\gO$, transition distribution $p_t(\vs_{h+1} \mid \vs_h, \va_h)$, reward $r_t$ and discount $\gamma$, where $h$ indexes timesteps within an episode.
The observation space $\gO$ and the continuous action space $\gA$ are shared across all tasks.
The goal is to maximise the expected discounted return
$J_i(\pi) = \E[\sum_h \gamma^h r_i(\vs_h, \va_h)]$ on each task $i \le t$ encountered so far, while only the current task's environment is available for the agent to interact with.
We solve each task using a parametric world model $\gW = (\vtheta, \vphi_t)$, where $\vtheta{\,\in\,}\sR^m$ denotes the task-agnostic parameters capturing reusable environment dynamics, and $\vphi_t{\,\in\,}\sR^n$ represents the task-specific parameters for each task.
The task-agnostic part compresses perception into a latent $\vz_h \in \gZ$ and predicts how that latent evolves, given very generally by
\vspace{-1pt}
\begin{equation}
  \vz_h \sim \ve_\vtheta(\cdot \mid \vz_{h-1}, \va_{h-1}, \vo_h),
  \quad
  \hat{\vz}_{h+1} \sim \vd_\vtheta(\cdot \mid \vz_h, \va_h),
  \label{eq:wm}
\end{equation}
\vspace{-1pt}
where $\ve_\vtheta$ encodes the current observation into the latent given the preceding latent and action, and $\vd_\vtheta$ predicts the next latent from the latent and action alone, enabling action-conditioned latent rollouts.
A set of heads $\vphi_t$ then consumes $\vz_h$ to produce the quantities needed for control on task $t$, e.g., reward, termination, value estimation, and a policy.
The backbones we evaluate populate the general description in Equation~\ref{eq:wm} differently.
DreamerV3~\cite{hafner2023mastering} carries the recurrence in $e_\theta$ through an explicit sequence model with a stochastic latent, reconstructs observations with a decoder, and acts through its policy directly, while TD-MPC2~\cite{hansen2024td} and PWM~\cite{li2025prismatic} produce a deterministic latent through a feed-forward network, are decoder-free, and use the policy as a prior for planning.
These differences change which components instantiate $(e_\theta, d_\theta, \phi_t)$, but not the fact that every model in this family is an action- and perception-conditioned latent predictor.

Our problem formulation exposes two properties that our benchmark builds on throughout: the two inputs a world model is conditioned on, and the separation of its parameters.

\paragraph{The world model input perspective (Section~\ref{sec:method-action-and-perception})}
A world model is characterised by its input modality at each step, namely \emph{action} and \emph{perception}. These modalities enter the model through different components: perception is consumed by the encoder before the dynamics are queried, while the action conditions the dynamics model directly. 
This perspective motivates us to factorise the composition task suites along these two axes, resolving a single compositional score into where reuse breaks.

\paragraph{The world model architecture perspective (Section~\ref{sec:method-task-agnostic-specific})} 
We partition a world model into the task-agnostic backbone $\theta$ and the task-specific heads $\phi_t$, e.g., state prediction is task-agnostic while reward prediction is task-specific.
This detail motivates how we apply continual learning to world models.

Section~\ref{sec:method-baseline-and-interface} adapts our baselines to the benchmark interface, and Section~\ref{sec:method-continual-learning} applies continual learning to monolithic and modular world models, with a frozen-encoder diagnostic that isolates what modularity contributes.

\subsection{Factorising composition along action and perception}
\label{sec:method-action-and-perception}
At each step, a world model takes two inputs: an \emph{action} to be taken, and current \emph{perception}, which in our case consists of multiple camera views and proprioception.
The dynamics it must learn are therefore conditioned on both.
This motivates the central design choice of our benchmark's task suites, we compose primitive skills along these two input modalities, yielding three task suite types:

\paragraph{Action-composition}
The perception context is held fixed.
The scene and objects stay consistent while the final composition task combines action primitives learnt through the curriculum, e.g., reaching and grasping.
This design isolates whether the model reuses across the action space under a familiar perception input.

\paragraph{Perception-composition}
The action primitives are held approximately fixed. The same manipulation is performed on objects with different visual appearances, and the composition task operates on an unseen combination of them.
This design isolates whether the model reuses across the perception space under a familiar action distribution.

\paragraph{Full composition}
Both action and perception axes change simultaneously. The primitive tasks differ in both action primitive and perception context, and the composition task combines both.

As discussed in Section~\ref{sec:relatedworks}, no existing benchmark ends a curriculum with a composition task that recombines primitives preceding it to isolate reuse, let alone factorises that composition by input modality.
We construct our benchmark using Meta-World's~\cite{yu2020meta} assets and environments, designing six composition task suites in total listed in Table~\ref{tab:task_suites}.

\begin{table}[t]
\vspace*{6pt}
\centering
\caption{\textbf{The six composition task suites.} primitives are trained in the listed order, final composition task is shown in \textbf{bold}}
\label{tab:task_suites}
\vspace{-5pt}
\scriptsize
\setlength{\tabcolsep}{3pt}
\renewcommand{\arraystretch}{1.15}
\begin{tabularx}{\linewidth}{@{}l X@{}}
\toprule
\textbf{Suite} & \textbf{Primitive Tasks} $\rightarrow$ \textbf{Composition task} \\
\midrule
\multicolumn{2}{@{}l}{\textit{Action composition}} \\
\addlinespace[2pt]
Reach       & reach in \textit{xy (plane)} $\rightarrow$ reach in \textit{xz} $\rightarrow$ reach in \textit{yz} $\rightarrow$ \textbf{reach in \textit{xyz}} \\
Grasp       & pick cube $\rightarrow$ place cube $\rightarrow$ \textbf{pick and place cube} \\
\midrule
\multicolumn{2}{@{}l}{\textit{Perception composition}} \\
\addlinespace[2pt]
BinPnP      & pick and place cube from red bin to blue bin \newline $\rightarrow$ yellow to blue $\rightarrow$ red to purple $\rightarrow$ \textbf{yellow to purple} \\
PnPBlock    & pick and place red block $\rightarrow$ pick and place blue block  \newline $\rightarrow$ \textbf{stack the two blocks} \\
\midrule
\multicolumn{2}{@{}l}{\textit{Full composition}} \\
\addlinespace[2pt]
DrawerPnP   & pick and place cube $\rightarrow$ open drawer \newline $\rightarrow$ \textbf{open drawer then pick and place cube inside} \\
PnPBoxClose & pick and place cube $\rightarrow$ close box $\rightarrow$ \newline \textbf{pick and place cube inside box then close it} \\
\bottomrule
\end{tabularx}
\vspace{-15pt}
\end{table}

\subsection{Model separation and continual learning adaptation}
\label{sec:method-task-agnostic-specific}
\begin{figure}[t]
    \centering
    \includegraphics[
        width=\linewidth,
        trim=35 100 25 12,
        clip
    ]{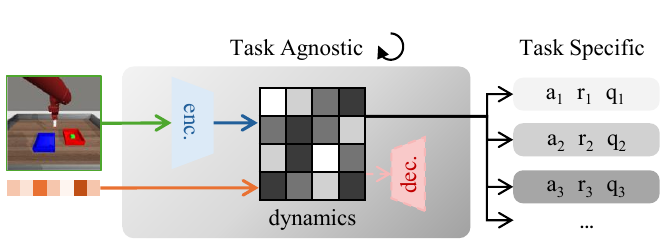}\
    \vspace{-15pt}
    \caption{\textbf{World model separation for continual learning.} The \emph{task-agnostic} part captures how the world evolves and is learnt continually across the curriculum. The \emph{task-specific} heads consume the dynamics latent and are re-initialised for each new task.}
    \label{fig:model-separation}
    \vspace{-15pt}
\end{figure}

Our second perspective is a structural way of applying continual learning in world models. As illustrated in Figure \ref{fig:model-separation}, a world model naturally partitions into two groups by semantic role. The \emph{task-agnostic} components capture how the world evolves and how observations are generated, and should be independent of any specific task. The \emph{task-specific} components consume the latent representation produced by the task-agnostic components to decide what the robot should do, and depend entirely on the task definition. 

This partition is consequential for how we apply continual learning, illustrated in Algorithm~\ref{alg:curriculum}.
When a new task arrives, the novel observations and dynamics must be absorbed by the task-agnostic part of the world model continually.
The task-specific heads are instantiated for each task and re-initialised when the curriculum advances.
This separation is because different tasks can be modelled using a shared world model, while each has its own reward and optimal policy.

\subsection{Baseline world models and interface adaptation}
\label{sec:method-baseline-and-interface}
We evaluate three world-model families under the model-based reinforcement learning view introduced above, namely DreamerV3~\cite{hafner2023mastering}, TD-MPC2~\cite{hansen2024td}, and prismatic world model (PWM)~\cite{li2025prismatic}. The first two serve as our monolithic baselines, while PWM is our modular baseline.

We present all baselines with the same observation space $\mathcal{O}$ and action space $\mathcal{A}$. Observations $o_t \in \mathcal{O}$ consist of channel-stacked images from three camera views (\emph{topview}, \emph{frontview}, \emph{gripperPOV}) and a $7$-D proprioceptive vector comprising end-effector pose, velocity, and gripper status.
The action space $a \in \mathbb{R}^4$ comprises relative end-effector displacement $\Delta p \in \mathbb{R}^3$ and a scalar gripper command $g \in \mathbb{R}$.
Each task has a success-based absorbing state.
We rescale the original reward of each task to $[-1, 0]$ to encourage minimum-time solutions when combined with the absorbing state.

Fitting each baseline to this interface requires three modifications: (i) widening or replacing the visual encoder, typically designed for a single low-resolution frame, so that it accepts the $9$-channel multi-camera image; (ii) adding an MLP branch for proprioception and a fusion step that combines the two modalities before forming the latent state; and (iii) ensuring that the action interface supports continuous control.
Beyond these changes, we preserve the native architecture, planning or imagination procedure, and actor-critic training of each baseline as faithfully as possible.

\begin{figure}[t!]
\setlength{\textfloatsep}{5pt}
\begin{minipage}{\linewidth}
\begin{algorithm}[H]
\caption{Continual learning and evaluation protocol}
\label{alg:curriculum}
\footnotesize
\begin{algorithmic}[1]
\Statex \textbf{Input:} $T$ tasks; CL method $\mathcal{C}$; World model $\mathcal{W}=(\theta, \phi)$
\State initialise task-agnostic backbone $\theta$
\For{$t = 1$ \textbf{to} $T$}
    \State initialise fresh task-specific heads $\phi_t$
    \While{task $t$ training}
        \State $\theta \gets \mathcal{C}(\theta)$ \Comment{update backbone continually}
        \State $\phi_t \gets \mathrm{update}(\phi_t)$ \Comment{update heads independently}
    \EndWhile
    \State store $\phi_t$ \Comment{freeze head for task $t$}
    \For{$i = 1$ \textbf{to} $t$}
        \State $R_i(t) \gets \mathrm{evaluate}(\theta, \phi_i)$ on task $i$ \Comment{eval task $i$}
    \EndFor
\EndFor
\State \Return $\{R_i(t)\}$
\end{algorithmic}
\end{algorithm}
\end{minipage}
\vspace{-15pt}
\end{figure}

\subsection{Continual learning with world models}
\label{sec:method-continual-learning}

\paragraph{Monolithic world model}
Following the model separation of Section \ref{sec:method-task-agnostic-specific}, we apply three popular continual learning methods to the task-agnostic parts, spanning the canonical families of replay, regularisation, and parameter isolation.
The implementation follows their original formulations~\cite{rolnick2019experience, kirkpatrick2017overcoming, mallya2018packnet} in DreamerV3 and TD-MPC2:
\textbf{Experience replay (ER)} represents the replay family.
We maintain a buffer holding $5\%$ ~\cite{rolnick2019experience} of each previously seen task's raw transitions and mix uniform samples from it into the backbone updates for the current task, so that earlier dynamics continue to be rehearsed as new tasks arrive.
\textbf{Elastic weight consolidation (EWC)}~\cite{kirkpatrick2017overcoming}
represents the regularisation family.
After each task, we estimate the diagonal Fisher information of the parameters to represent the importance of each parameter, and add a quadratic penalty that discourages important ones from drifting.
\textbf{PackNet}~\cite{mallya2018packnet}
represents the parameter-isolation family.
After training on each task, we prune the backbone to retain $25\%$~\cite{mallya2018packnet} of its remaining free parameters and freeze that subset as a protected sub-network.
Together with naive \textbf{Fine-tuning (FT)}, four continual learning paradigms are adapted for each monolithic world model to benchmark.
\begin{figure*}[t!]
    \centering
    \includegraphics[trim={200pt 65pt 195pt 70pt}, clip, width=\textwidth]
    {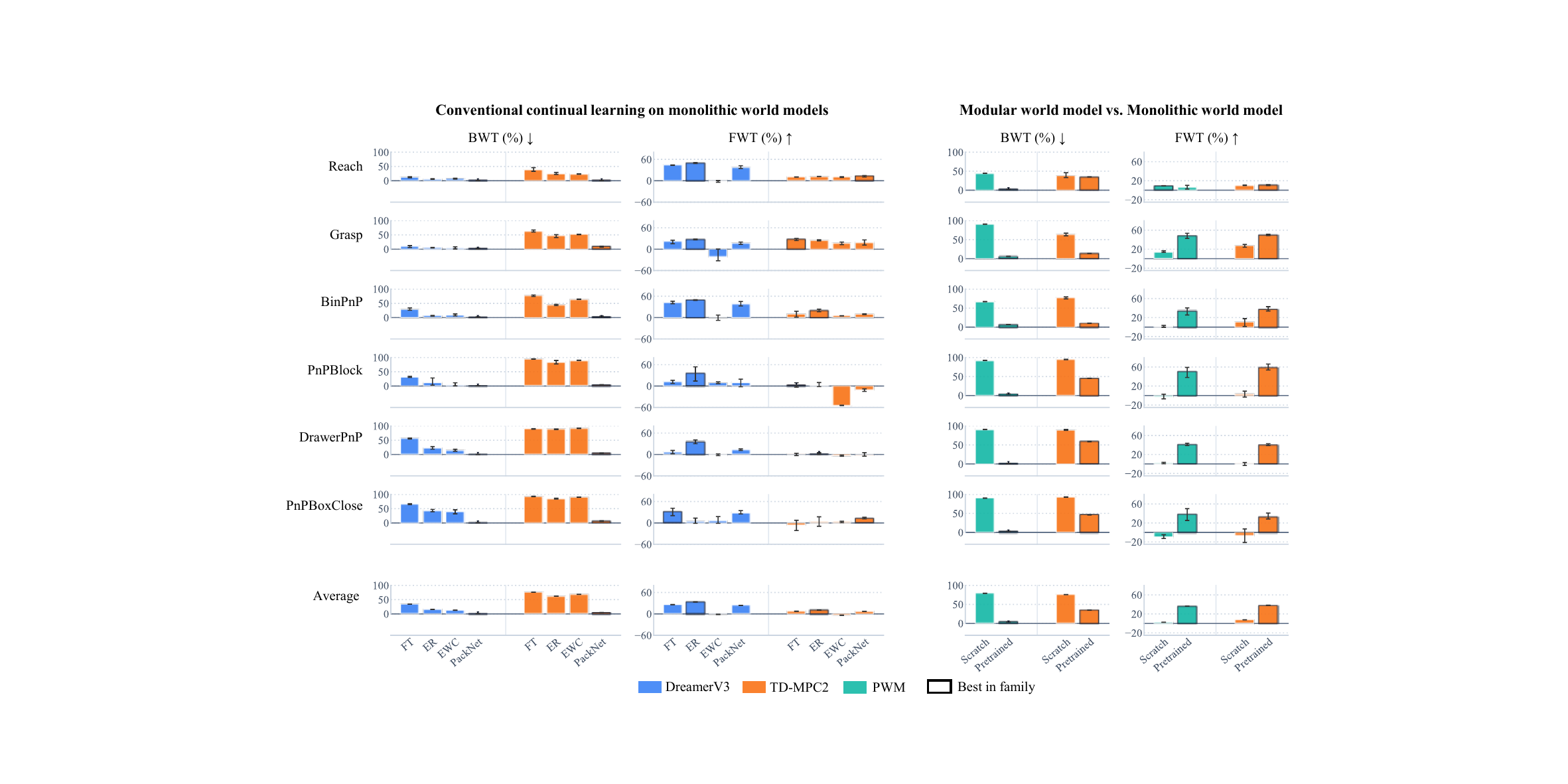}
    \vspace{-10pt}
    \caption{\textbf{Continual learning performance on all six task suites.} Task suites are shaded by composition type. Lower BWT and higher FWT are better, black outlines mark the best method within a backbone. \textit{Left}: the monolithic backbones DreamerV3 (blue) and TD-MPC2 (orange) under four continual learning methods. \textit{Right}: modular PWM (teal) against monolithic TD-MPC2 (orange), with a scratch or a pretrained frozen encoder. Bars show the mean over three random seeds with minimum and maximum scores indicated.}
    \label{fig:results-bars}
    \vspace{-15pt}
\end{figure*}

\paragraph{Modular world model}
For the modular world model motivated in Secion \ref{sec:introduction}, we build upon the mixture-of-experts (MoE) world model PWM~\cite{li2025prismatic}, which extends TD-MPC2 to multiple dynamics experts that produce intermediate representations, combined by a learnt soft router to produce the final latent, making it a close counterpart to TD-MPC2.

The original PWM, however, is designed for multi-task learning.
It trains on all tasks jointly and provides no mechanism for tasks that arrive in sequence, so we first construct a continual variant by transferring the progressive mixture-of-experts idea from Wang et al.\cite{wang2024sparse}.
For each incoming task we introduce $K=3$ new trainable dynamics experts, while the experts of previous tasks are frozen. The router can access all experts activated up to the current task, so the model can reuse previously learnt dynamics while still introducing new capacity when needed.

\paragraph{Frozen-encoder diagnostic}
While our modification lets PWM's dynamics model learn and combine experts progressively along the curriculum, the results it produces do not yet isolate what the modular design contributes.
Its encoder remains a single network that keeps updating throughout the curriculum. The experts therefore operate on a latent representation that is itself drifting, and any result is thus entangled with this encoder drift.

To remove this effect and isolate what the modular architecture contributes, we pretrain the encoder in PWM with a simple autoencoder reconstruction objective on expert demonstrations of all tasks in our benchmark, then freeze it throughout sequential learning.
To control for the contribution of this privileged encoder, we use the exact same encoder checkpoint on the monolithic TD-MPC2 for comparison, attributing any remaining performance gap to the modular dynamics design, up to the additional capacity it introduces.

\section{Experiments}
\label{sec:result}

Following the continual learning protocol discussed in Section \ref{sec:method-task-agnostic-specific}, we report two metrics:

\textbf{Backward transfer (BWT)}~\cite{lopez2017gradient} measures how much earlier tasks degrade after their training ends and new tasks start.
Lower BWT indicates less forgetting.

\textbf{Forward transfer (FWT)}~\cite{wolczyk2021continual} measures how fast the final composition task is learnt relative to learning it from scratch.
Higher FWT indicates more reuse.
FWT is only measured on the final composition task to evaluate reusing.

Cumulative discounted returns achieved during training and evaluation are normalised between a failure floor of $0$ and a scripted oracle policy reference of $100$, further enabling FWT and BWT calculation.
We first report the BWT and FWT scores of different backbones and methods under our benchmark (Figure~\ref{fig:results-bars}) and analyse the results based on our composition point of view in Section~\ref{sec:main-results}.
Then we investigate whether the modular world model actually reuses its previously learnt components through visualisations and ablations in Section~\ref{sec:router-and-ablation}.
Further details on metric definitions, per-task absolute returns for each backbone, full numerical results, and complete training curves are provided in the appendix section on our \href{https://object814.github.io/Compositional-Continual-Learning/}{project website}.

\vspace{-5pt}
\subsection{Baseline Performance and Comparison}
\label{sec:main-results}

\paragraph{DreamerV3 outperforms TD-MPC2}
Under naive fine-tuning and all three conventional continual learning methods, DreamerV3 consistently outperforms TD-MPC2 on both BWT and FWT, as shown in the left half of Figure~\ref{fig:results-bars}.
We attribute this to DreamerV3's generative reconstruction objective, which encourages its latent to encode the scene more broadly than the current task alone requires.
In contrast, TD-MPC2 is decoder-free by design and shapes its latent only through latent consistency, reward prediction and TDlearning, none of which requires the representation to retain scene detail the current task does not use.

\paragraph{Conventional methods trade forward transfer against backward transfer} No conventional method achieves both, each sits at a different point along the same trade-off.

\textit{Fine-tuning} is second only to ER in forward transfer. When the composition task arrives, the backbone already encodes the required primitives, and every parameter is directly available. But with no forgetting prevention, the parameters it reads are the ones it writes, so nothing separates recombining a mechanism from overwriting it, giving the most forgetting.

\textit{ER} preserves what fine-tuning reuses, keeping predictions on earlier dynamics correct while parameters still move freely to recombine for the composition task. This yields the best forward transfer of all, with roughly half fine-tuning's forgetting. However, drift is only slowed rather than stopped, and the buffer grows unbounded as more tasks arrive.

\textit{EWC} prevents forgetting through penalising updates to the parameters holding prior knowledge. It can then neither move those parameters to recombine them, as fine-tuning and ER do, nor fall back on free capacity to reuse them, as PackNet does. It is the weakest method here, retaining less well than PackNet while leaving the model worse at learning the composition task than starting from scratch.

\textit{PackNet} comes closest because freezing is paired with a pathway for reusing.
While pruning and freezing one sub-network per task, task $t$ during inference activates every parameter allocated to tasks $n \leq t$, so primitives remain available as reusable components, and the pruned remainder is unpenalised from learning a combination over previous knowledge.
PackNet therefore drives forgetting near zero while retaining competitive forward transfer.
The escape is only partial, as model capacity shrinks along the sequence, so forward transfer still trails fine-tuning and ER.
Also, pruning allocates a fixed fraction per task and bounds curriculum length in advance, making PackNet the least scalable.

\paragraph{Explicit modularity pushes the frontier further}
PWM preserves reusable components and enables recombination through an explicit architectural structure in the dynamics model.
Each task adds its own dynamics experts, which we fix at three, while previous ones are frozen, and a router learns to combine over them. Comparing it against its direct monolithic counterpart TD-MPC2 under our frozen encoder diagnostic to isolate modular design contributions, we see a sharp improvement, shown in the right half of Figure~\ref{fig:results-bars}.
Averaged across the six task suites, PWM reaches BWT $3.85$ and FWT $36.18$ against TD-MPC2's $34.97$ and $38.11$, nearly eliminating forgetting by freezing experts while matching how fast TD-MPC2 learns the composition task under fine-tuning, a stronger baseline for forward transfer.
Adaptation survives because the router supplies combining capacity, and unlike PackNet, this capacity is added with each task rather than carved from a shrinking pool, making the method more scalable.
However, modularity does not push the reuse axis further, as forward transfer only matches what the monolithic model already achieved. Moreover, PWM's advantage holds only under the privileged frozen encoder, without which its performance is on par with TD-MPC2.
Therefore, a design that reuses components more effectively, and an encoder that supplies a stable representation without privileged access to the full task distribution, both remain open.

\begin{table}[t]
\vspace*{6pt}
\centering
\caption{\textbf{Performance by composition axis.} BWT$\downarrow$ and FWT$\uparrow$ grouped by composition type, each averaged over all four continual learning methods and both task suites of that modality. \textbf{Bold} indicates the best performance of each backbone across the composition axis.}
\label{tab:modality}
\vspace{-5pt}
\footnotesize
\renewcommand{\arraystretch}{1.05}
\begin{tabular*}{\linewidth}{@{\extracolsep{\fill}} l cc cc cc @{}}
\toprule
 & \multicolumn{2}{c}{Action} & \multicolumn{2}{c}{Perception} & \multicolumn{2}{c}{Full} \\
\cmidrule(lr){2-3}\cmidrule(lr){4-5}\cmidrule(lr){6-7}
Backbone & BWT$\downarrow$ & FWT$\uparrow$ & BWT$\downarrow$ & FWT$\uparrow$ & BWT$\downarrow$ & FWT$\uparrow$ \\
\midrule
DreamerV3 & \textbf{5.94} & $21.56$ & $11.79$ & \textbf{24.52} & $30.12$ & $15.54$ \\
TD-MPC2 & \textbf{32.15} & \textbf{16.76} & $57.81$ & $-1.88$ & $68.68$ & $1.54$ \\
\bottomrule
\end{tabular*}
\vspace{-18pt}
\end{table}

\paragraph{Difficulty rises along the composition axis}
We average the two suites of each composition type over all four continual learning methods, to represent the overall performance on that composition axis, in Table~\ref{tab:modality}. 
BWT worsens monotonically from action to perception to full composition on both backbones. 
FWT instead separates them. DreamerV3 stays within a narrow band across all three axes, whereas TD-MPC2 loses forward transfer entirely once perception is involved. This ordering follows where each axis enters the model.
Recombining actions under a familiar scene leaves the dynamics model a shifted action distribution over a stable visual latent, whereas a perception shift lands on the encoder and also reaches the dynamics model through a drifting visual latent.
Full composition perturbs both inputs directly at once, and neither backbone reuses
without forgetting there.

\subsection{Probing Reuse in the Modular World Model}
\label{sec:router-and-ablation}

\begin{figure*}[t!]
\centering
\includegraphics[trim={225pt 150pt 350pt 110pt}, clip, width=\textwidth]{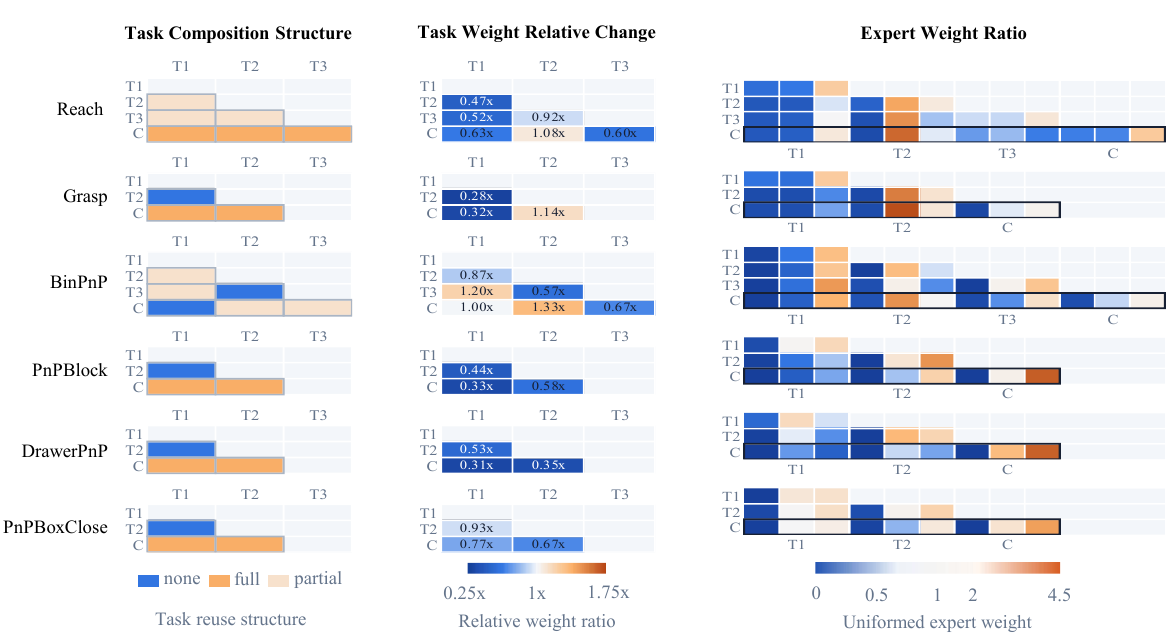}
\vspace{-15pt}
\caption{\textbf{Task composition structure and expert routing dynamics.}
\textit{Left}: Task-composition structure for each curriculum. For a matrix with rows $i$ and columns $j$, entry $(i,j)$ indicates whether task $i$ contains a reusable component from task $j$ along the corresponding action, perception, or full-composition axis.
\textit{Middle}: Relative change in the average routing weight assigned to the three experts introduced by each task. Each entry is expressed as a ratio to that task's weight immediately after learning the corresponding task itself; $1\times$ denotes no change, values below $1\times$ indicate down-weighting, and values above $1\times$ indicate increased weighting. If PWM reuses experts according to the task-composition structure, the pattern in this panel should broadly correspond to that on the \textit{left}.
\textit{Right}: Detailed routing weights over individual experts at each curriculum stage, normalised by the uniform allocation $1/K_{\mathrm{active}}$, such that $1.0$ denotes uniform weighting. Each row shows the expert weight allocation after training on a task, while each column tracks the routing weight of a specific expert over the curriculum. The boxed final row shows routing after learning the composition task.}
\label{fig:router-visualisation}
\vspace{-18pt}
\end{figure*}

Section \ref{sec:main-results} showed that, under a frozen encoder, PWM largely eliminates forgetting while retaining competitive forward transfer. However, it remains less clear whether adaptation is based on reusing previous experts or on learning the newly introduced ones. Our benchmark's explicit compositional structure allows us to examine this distinction directly. We therefore probe the pretrained, frozen-encoder PWM by first inspecting whether routing reflects the known task composition, then ablating expert access and routing to test how prior experts, new experts, and learnt weighting contribute to composition-task performance.

\paragraph{Router weights recover task-level reuse structure in several curricula}
Fig.~\ref{fig:router-visualisation} compares the compositional structure between tasks with how PWM reallocates weight to previously introduced experts. Across several curricula, the learnt routing broadly follows the expected reuse pattern, particularly for the \texttt{Reach}, \texttt{Grasp}, and \texttt{BinPnP} suites. Experts associated with recurring components tend to retain more of their original weight, while unrelated experts are more strongly down-weighted. \texttt{BinPnP} provides a clear example during primitive-task learning: on the third task, the router increases the weight assigned to experts from an overlapping prior task to $1.20\times$ while reducing that of a non-overlapping one to $0.57\times$. Similar trends in \texttt{Grasp} and \texttt{Reach} also suggest that PWM selectively allocates weight to prior experts when their underlying components recur.

\paragraph{Task-level routing may limit temporal composition}
However, this correspondence between routing and task structure is not consistent across all task suites.
In particular, we do not observe the expected reuse pattern in \texttt{PnPBlock} and the two full-composition suites after composition learning.
We attribute this to PWM's router being task-conditioned rather than state-conditioned.
A fixed expert allocation is learnt for each task, preventing the router from varying its allocation on-the-fly.
This fixed router may limit its suitability for temporally composed tasks, where different primitives are required at different stages.
Notably, the three suites without a clear routing correspondence involve such temporal composition.
More generally, composition tasks may also introduce interactions that are not captured by either primitive in isolation, such as transitions between them. 
These observations suggest that effective reuse may require both routing that adapts across stages of a task and additional capacity to capture interactions between primitives.

\paragraph{Composition learning preserves expert-level preferences}
The right panel of Figure\ref{fig:router-visualisation} looks closer into the weight distribution for individual experts, and shows that expert-level preferences persist across all learning curriculums. The highest-weighted expert within each primitive task remains top-ranked after composition learning in $13$ out of $14$ cases.
Moreover, weight remains distributed across prior and new experts rather than concentrating exclusively on the newly introduced capacity.
This observation motivates us to design the following ablation to test whether their routing weights correspond to a functional contribution.

\begin{figure}[t!]
    \centering
    \includegraphics[trim={410pt 140pt 450pt 110pt}, clip, width=0.95\linewidth]{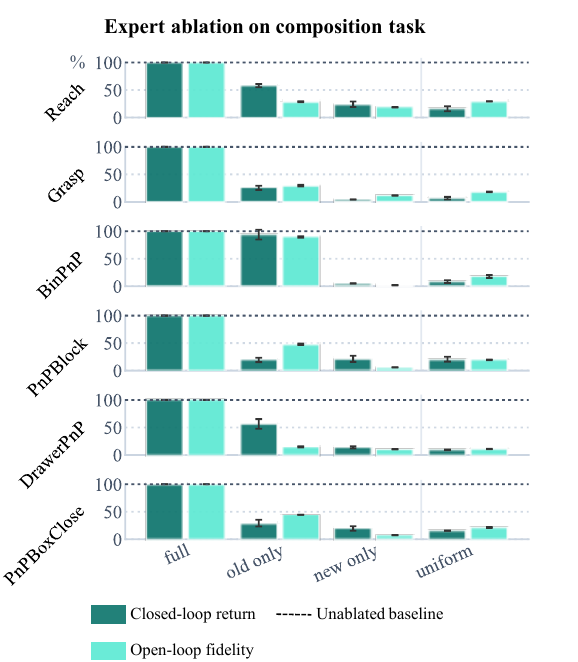}
    \vspace{-5pt}
    \caption{\textbf{Expert ablation on the composition task.} Closed-loop return and open-loop latent prediction fidelity on the composition tasks when activating only prior experts, only new composition task experts, or uniformly weighting all, relative to the full learnt mixture. Prior experts retain substantial performance, while new experts alone and uniform routing perform worse.}
    \label{fig:expert-ablation}
    \vspace{-18pt}
\end{figure}

\paragraph{Expert ablation confirms functional reuse of prior experts}
To test whether routing patterns reflect functional reuse, Figure \ref{fig:expert-ablation} compares closed-loop return and open-loop latent prediction fidelity under three ablations: retaining only prior experts from primitive tasks, retaining only new experts for composition task, and uniformly weighting all, each relative to the full learnt mixture. Prior experts alone retain near-baseline performance on \texttt{BinPnP}, providing clear functional support for our routing analysis. They also preserve substantial return on \texttt{Reach} and \texttt{DrawerPnP}, although prediction fidelity drops more sharply. The \texttt{DrawerPnP} result demonstrates that useful prior dynamics remain available even when routing does not clearly reflect expected compositional structure. New experts alone perform poorly across suites, while uniform routing falls well below the full mixture, showing that having experts is insufficient without learning their routing.
Together, these results are consistent with functional reuse of prior dynamics through learnt routing.

\section{Conclusion}
\label{sec:conclusion}

We propose a compositional continual learning benchmark for world models in robot manipulation, where each curriculum ends with a task built by recombining the primitives preceding it, so that forward transfer isolates reuse from raw learning speed. We also factorise the composition along action and perception to localise where reuse without forgetting fails.
Furthermore, we introduce a principled separation of world models into a continually learnt task-agnostic backbone and task-specific heads, giving a consistent way to apply continual learning across this model family.

Our evaluation shows that no conventional continual learning method solves reuse without forgetting, all trading one side of it for the other, and this difficulty increases along the action, perception, and full composition axes.
An explicit modular dynamics model drives forgetting close to zero while matching the forward transfer of its monolithic counterpart, but only under a diagnostic encoder pretrained on demonstrations of all tasks and then frozen.
Without this privilege, it is on par with that counterpart throughout.
We read this as evidence for what explicit reusable model design can buy, rather than as a state-of-the-art result.
These results motivate world models that can combine prior experts more effectively, and leverage general pre-trained features such as the DINOv2 features used by DINO-WM~\cite{zhou2024dino}, towards a fully task-agnostic backbone that reuses without forgetting.

We acknowledge several limitations of our benchmark that can be improved. Our evaluation is limited to simulated curricula, online MBRL world models, and a single model size or expert number per baseline without controlled parameter or compute budgets.
We omit real-robot deployment because online MBRL requires on the order of $10^{5}$--$10^{6}$ environment steps per task, repeated across six suites, multiple backbones, methods and seeds, and because our protocol relies on dense rewards, oracle-normalised returns and automatic resets that are simulator affordances. Porting the compositional protocol to hardware therefore depends on offline-pretrained backbones and learnt success detection, which we leave to future work.


\newpage
\bibliographystyle{IEEEtran}
\bibliography{reference}

\end{document}